\documentclass[conference]{IEEEtran}
\IEEEoverridecommandlockouts
\usepackage{cite}
\usepackage{amsmath,amssymb,amsfonts}
\usepackage{algorithmic}
\usepackage{graphicx}
\usepackage{textcomp}

\usepackage[table,xcdraw]{xcolor}

\usepackage{subfig}

\usepackage{multirow}
\usepackage{float}

\def\BibTeX{{\rm B\kern-.05em{\sc i\kern-.025em b}\kern-.08em
    T\kern-.1667em\lower.7ex\hbox{E}\kern-.125emX}}
\begin{document}

\title{Bayesian Federated Learning for Continual Training\\
\thanks{Funded by the European Union (EU). Views and opinions expressed are however those of the author(s) only and do not necessarily reflect those of the EU or European Innovation Council and SMEs Executive Agency. Neither the EU nor the granting authority can be held responsible for them. Grant Agreement No: 101099491 (Holden project). The work is also supported by CNR-MOST (2024/2025) bilateral collaboration (SAC.AD002.184).}
}
\author{\IEEEauthorblockN{Usevalad Milasheuski\textit{$^{1,2}$}, Luca Barbieri\textit{$^{3}$},
		Sanaz Kianoush\textit{$^{2}$}, Monica Nicoli\textit{$^{1}$}, Stefano Savazzi\textit{$^{2}$}} \IEEEauthorblockA{\textit{$^{1}$}\textit{\emph{ }}DEIB and DIG, Politecnico di Milano, Piazza
		Leonardo da Vinci 32, I-20133, Milan, Italy\\
		\textit{$^{2}$}\textit{\emph{ Consiglio Nazionale delle Ricerche}}\emph{,}
		\textit{\emph{IEIIT}} institute, Piazza Leonardo da Vinci 32, I-20133,
		Milan, Italy.\\
		\textit{$^{3}$}\textit{\emph{ Nokia Bell Labs, Magirusstraße 8, 70469, Stuttgart, Germany}}
}}

\maketitle

\begin{abstract}
Bayesian Federated Learning (BFL) enables uncertainty quantification and robust adaptation in distributed learning. In contrast to the frequentist approach, it estimates the posterior distribution of a global model, offering insights into model reliability. However, current BFL methods neglect continual learning challenges in dynamic environments where data distributions shift over time. We propose a continual BFL framework applied to human sensing with radar data collected over several days. Using Stochastic Gradient Langevin Dynamics (SGLD), our approach sequentially updates the model, leveraging past posteriors to construct the prior for the new tasks. We assess the accuracy, the expected calibration error (ECE) and the convergence speed of our approach against several baselines. Results highlight the effectiveness of continual Bayesian updates in preserving knowledge and adapting to evolving data.


\end{abstract}


\section{Introduction}
Federated Learning (FL) \cite{FL_main} is a privacy-preserving machine learning (ML) paradigm that enables multiple parties to collaboratively train a global model without sharing their raw data. This distributed training addresses critical security and data ownership concerns, making it useful when sensitive data is prevalent. By distributing the computation, FL mitigates data transmission risks while enabling large-scale learning across diverse datasets. Its versatility is widely adopted in healthcare \cite{FL_health}, smart homes \cite{FL_snarthome}, autonomous vehicles \cite{FL_auton}, and industry \cite{FL_telco,comm_mag}. However, its frequentist nature provides only a single-point estimate, limiting uncertainty representation and increasing overfitting risks.

\begin{figure} 
\centerline{\includegraphics[width=0.5\textwidth]{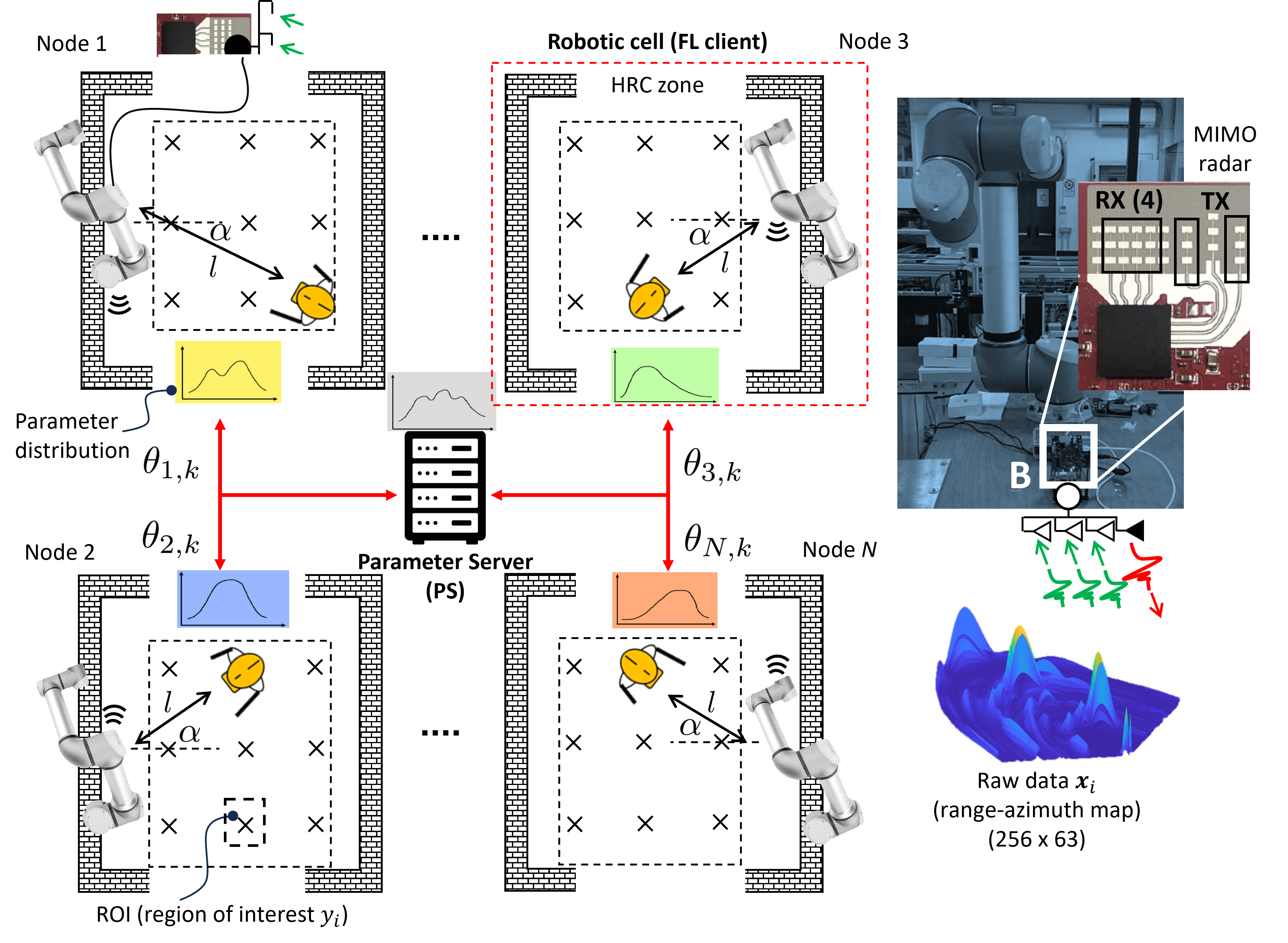}}
    \caption{Bayesian FL for human sensing. Each robot cell is equipped with a radar measuring the angle $\alpha$ and the distance $l$ to the operator (range-azimuth map).} 
    \label{fig:bfl}
\vspace{-0.2cm}    
\end{figure} 
To address these limitations, Bayesian Federated Learning (BFL) \cite{BFL} applies Bayesian inference \cite{BI} over the parameter space by maintaining a \textit{posterior distribution} which captures the inherent uncertainty in the model and data. This probabilistic perspective provides more robust predictions and improves generalization. When the posterior is intractable, two approaches can be used.  \textit{Variational Inference} (VI) techniques \cite{VI} approximate the posterior with simpler, tractable distributions by reformulating the problem as an optimization task which minimizes the Kullback-Leibler (KL) divergence between the approximation and the true posterior \cite{VI2}. \textit{Markov Chain Monte Carlo} (MCMC) techniques \cite{MCMC} approximate the posterior with a stationary distribution of Markov Chain (MC), allowing it to generate samples that approximate the true posterior. A widely used MCMC-based method is Stochastic Gradient Langevin Dynamics (SGLD) \cite{SGLD}, which combines Stochastic Gradient Descent (SGD) with Gaussian noise to simulate the dynamics of the posterior distribution. 
Various MCMC methods are studied in \cite{wirelessSGLD, MCMC3, MCMC4}, where the authors adopted a centralized FL approach by exchanging local posterior distribution samples produced by the participants with the Parameter Server (PS).

In scenarios where data is collected sequentially, such as medical monitoring \cite{cont_health} or Industrial Internet of Things (IIoT) processes \cite{cont_iot}, \textit{Continual Learning} (CL) \cite{CL} becomes essential for adapting to new information without discarding previously acquired knowledge. Traditional FL approaches often struggle to address evolving data distributions. Despite the lack of the analysis of BFL for CL \cite{CBFL1, CBFL2}, with its ability to incorporate meaningful prior, it provides a natural solution for continual learning. 

\textbf{Contributions:} we explore a continual BFL framework based on SGLD. The framework is verified in an industrial IoT use case, where a model is trained to assist operators in monitoring a human-robot collaborative work environment. The model is subject to periodic and continual adaptation based on the new data coming from the following days. The goal is to investigate two key research objectives: 
\begin{enumerate}
    \item 
     evaluate the benefits of adapting posterior distributions as the next priors for CL in a sequential way, which we refer to as Posterior-aided Continual Learning (P-CL);  
    \item compare the performance of the proposed sequential adaptation with the conventional CL paradigms.
\end{enumerate}

This paper is organized as follows. Sec. \ref{BFL} provides the overview of BFL and SGLD. Sec. \ref{CBFL} describes the analyzed continual BFL framework. Sec. \ref{radar_and_problem} describes the industrial case study used for the evaluation of our continual BFL framework. Sec. \ref{RES} conducts the overview of the results, while Sec. \ref{CONCL} discusses the conclusion and future work.

\section{Distributed Bayesian Optimization} \label{BFL}

In this work, we focus on the conventional synchronous Parameter Server (PS) based FL setup \cite{FL_main}, as illustrated in Fig. \ref{fig:bfl}. We consider a set of $N$ nodes, where each node $n$ has a local dataset $\mathcal{D}_n = \{(\mathbf{x}_{n, i}, y_{n, i}) \}_{i=1}^{|\mathcal{D}_n|}$, $\mathbf{x}_{n, i}$ and $ y_{n, i}$ are inputs to the model and labels respectively,  $\mathcal{D} = \{\mathcal{D}_n\}_{n=1}^N$. The training process follows an iterative scheme where each node member of the federation $n \in \{1, \dots, N\}$ performs local updates on its private dataset $\mathcal{D}_n$ and shares only model parameters or gradients with the PS, as shown in Fig. \ref{fig:bfl}. The goal of the devices is to collaboratively train a global model under the coordination of the PS, without sharing their local datasets

Bayesian Federated Learning (BFL) \cite{BFL} extends conventional FL by incorporating Bayesian inference to model uncertainty and improve generalization. Each node performs local Bayesian updates and contributes to the computation of global posterior distribution. Let $\boldsymbol{\theta} \in \mathbb{R}^P$ be the model parameters. Given a dataset $\mathcal{D}$, the posterior distribution of $\boldsymbol{\theta}$ is computed using the Bayes' theorem:

\begin{equation} \label{eq:bayes_theorem}
   p(\boldsymbol{\theta} | \mathcal{D}) \propto p(\boldsymbol{\theta})  \prod_{n=1}^{N} p(\mathcal{D}_n | \boldsymbol{\theta}),
\end{equation}
where $p(\mathcal{D}_n | \boldsymbol{\theta}) = \prod_{i=1}^{|\mathcal{D}_n|}p(y_{n, i}|\mathbf{x}_{n, i}, \boldsymbol{\theta})$is the likelihood of the data given the parameters for each node, $p(\boldsymbol{\theta})$ is the prior distribution over parameters.

A common approach to approximate these distributions is using gradient-based Markov Chain Monte Carlo (MCMC) methods, such as Stochastic Gradient Langevin Dynamics (SGLD), allowing efficient posterior sampling. This iterative approach can obtain a sample-based accurate approximation of the posterior by generating samples using a noisy version of SGD starting from the initial set of parameters $\boldsymbol{\theta}_{0}$

\begin{equation} \label{eq:sampling_1}
    \boldsymbol{\theta}_{k+1} = \boldsymbol{\theta}_{k} - \eta \nabla L(\boldsymbol{\theta}_{k}) + \sqrt{2 \eta} 
    \xi_{k+1},
\end{equation}
where $\eta$ is the learning rate, $\xi_k \sim \mathcal{N}(\mathbf{0}_{N_p}, \text{I}_{N_p})$ - vector drawn from a standard normal distribution, $L(\boldsymbol{\theta}_k) = \sum_{n=1}^{N}L_n(\boldsymbol{\theta}_k),$ 

\begin{equation}
    L_n(\boldsymbol{\theta}_k) = -\log p(\mathcal{D}_n | \boldsymbol{\theta}_k) -\frac{1}{N} \log p( \boldsymbol{\theta}_k).
\end{equation}
Typically, the sampling procedure should start once the MC has converged to a stationary distribution. Therefore, following \cite{SGLD}, a \textit{"burn-in" phase} is introduced, during which we discard the first $T_b$ samples, while the following $T_s=T-T_b$ samples are used for the posterior approximation.

\section{Continual Bayesian FL Methodology}\label{CBFL}
The MCMC-based BFL framework \cite{Luca} serves as an accurate decentralized method for the estimation of the posterior of the parameters. However, considering the nature of the FL process, the data coming from the nodes can not only be heterogeneous, but also dynamic. Thus, the posterior estimate at one point in time might become inaccurate as the data evolves. A solution could be to continually retrain the model on the newly collected data to adapt to the changes. While this solution is effective, it is inefficient in terms of the communication overhead. When the dynamics of the data are not abrupt, we could incorporate this knowledge into the training procedure.

This idea is aligned with BFL which relies on meaningful prior $p(\boldsymbol{\theta})$. We denote the dataset on node $n$ during day $d$ as $\mathcal{D}_n^d = \{(\mathbf{x}_{n, i}^d, y_{n, i}^{d}) \}_{i=1}^{|\mathcal{D}_n^d|}$. Considering the day $d$ and node $n$, the model parameters are updated as follows:

 \begin{equation}\label{eq:samp}
    \boldsymbol{\theta}_{n, k+1}^{d} = \boldsymbol{\theta}_{n, k}^{d} - \eta \nabla L(\boldsymbol{\theta}_{n, k}^{d}) + \sqrt{2 \eta} \xi_{k+1, n}, 
\end{equation}
where the gradient is computed as 
\begin{equation}
    \begin{aligned}
        \nabla L(\boldsymbol{\theta}_{n, k}^{d}) = & - \frac{1}{M} \sum_{m=1}^{M} 
        \sum_{\mathcal{M}_n^{d, (m)}\sim \mathcal{D}_n^d} 
        \nabla \log p(\mathcal{M}_n^{d, (m)} | \boldsymbol{\theta}_{n, k}^{d}) \\
        & -\frac{1}{N} \nabla \log p( \boldsymbol{\theta}_{n, k}^{d})
    \end{aligned}
\end{equation}
by sampling $M$ batches $\mathcal{M}_n^{d, (m)}$ from $\mathcal{D}_n^d$.

After a sufficient number of samples from (\ref{eq:samp}) is collected, we can reuse this information further. Given the dataset for the following day, the previous posterior estimate may be inaccurate for this dataset. If no significant changes in the system happened between $d$ and $d-1$, the information from collected posterior for $d-1$ may still be relevant for $d$. We propose to use the posterior $p(\boldsymbol{\theta} | \mathcal{D}_n^{d-1})$, as defined in (\ref{eq:bayes_theorem}), to compute the required statistic for the prior distribution of the following day. For instance, assuming the prior follows Gaussian distribution, we define the prior as
\begin{equation} \label{eq:bayes}
    p(\boldsymbol{\theta}^{d}) \sim \mathcal{N}(\boldsymbol{\mu}( \boldsymbol{\theta}^{d-1}), \mathbf{\Sigma}(\boldsymbol{\theta}^{d-1})),
\end{equation}
where $\mu( \boldsymbol{\theta}^{d})$ and $\mathbf{\Sigma}(\boldsymbol{\theta}^{d})$ are the mean and covariance matrix computed based on the samples collected during day $d$.

\begin{figure} 
\centerline{\includegraphics[width=0.48\textwidth]{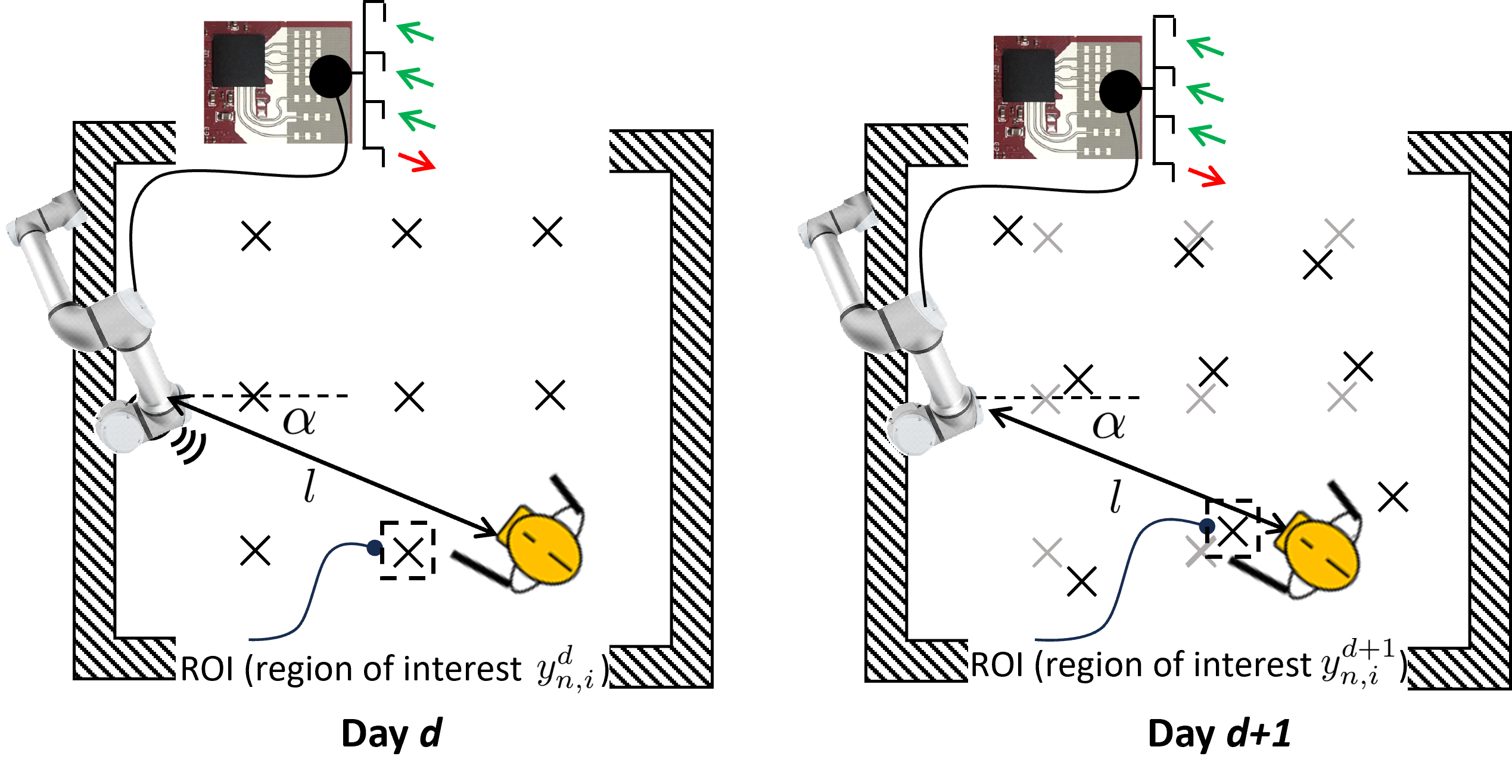}}
    \caption{Continual learning example. Black crosses represent the Regions of Interest (ROIs) for the current day, while the \textcolor{gray}{gray} ones represent the targets for the past day to highlight the misalignment between days.}
\label{fig:cont_learn_fig}
\vspace{-0.4cm}
\end{figure}

\section{Continual learning use case: human sensing} \label{radar_and_problem}
In the proposed FL setup, we use a network of sensors with a Time-Division Multiple-Input-Multiple-Output (TD-MIMO) Frequency Modulated Continuous Wave (FMCW) radars operating in the $77$-$81$ GHz band. The data contain raw range-azimuth measurements from mmWave MIMO radars deployed in a Human-Robot (HR) shared workspace environment, enabling real-time tracking of human operators. To ensure safety, the system continuously updates model parameters to adapt to changes in the workspace (i.e., minor modifications in the operating environment or workflow, as well as any shifts in the robot’s and operators' relative positions), as shown in Fig. \ref{fig:cont_learn_fig}, ensuring both safety and reliability. Each radar has an antenna array consisting of three transmitters (TX) and four receivers (RX), providing an azimuth Field of View (FOV) spanning $\pm 60^\circ$, an angular resolution of $25^\circ$, and a range resolution of 4.2 cm across a 3.9 GHz bandwidth with a 36 $\mu s$ sweep time. It monitors the distance ($l$) (i.e., range) and Direction of Arrival (DOA) ($\alpha$) (i.e., azimuth) of individuals relative to robotic manipulators within a workspace \cite{HR-INT}. Radars process the signals from reflections caused by moving objects via Fast Fourier Transform (FFT) to generate range-azimuth maps ($\mathbf{x}_i$) of size $256 \times 63$. The model classifies human presence within $R = 10$ predefined Regions of Interest (ROIs), representing different HR interaction scenarios. Specifically, ROI $y_i = 0$ corresponds to a human worker maintaining a safe distance ($\geq 2$ m), while $y_i > 0$ indicates closer proximity to the robotic system, with varying HR distances and DOA values.
To optimize model training without data transfer, a FL approach is employed, which enables distributed model training while maintaining data privacy. We resort to a Bayesian version of FL to estimate the posterior distribution of the model parameters, which helps estimate how confident the acquired model is in its predictions for uncertainty quantification, which is vital to ensure human safety  \cite{RH-SAFETY}.
The proposed continual BFL setup was evaluated using a simulated environment, where virtual radar devices process assigned range-azimuth data and ROI \cite{dataset} and exchange the updates, as described in Section III. We adopt a LeNet architecture \cite{LeNet} with approximately 1.4 million parameters. The model is subject to adaptation to accommodate dynamic changes in the operational environment. The initial model training occurs on the day $d = 1$, with a dataset 
$\mathcal{D}^1 = \{(\mathbf{x}_{i}^1, y_{i}^1) \}_{i=1}^{|\mathcal{D}^1|}$ with a pre-defined prior $p(\boldsymbol{\theta}^{1})$, while for the following days $d>1$ the model is trained on the new data with prior according to (\ref{eq:bayes}). 

\section{Accuracy vs. Confidence results and analysis} \label{RES}

\captionsetup[subfigure]{labelformat=empty}
\begin{figure*}[!ht]
    \centering
    \subfloat{
        \includegraphics[width=0.23\textwidth]{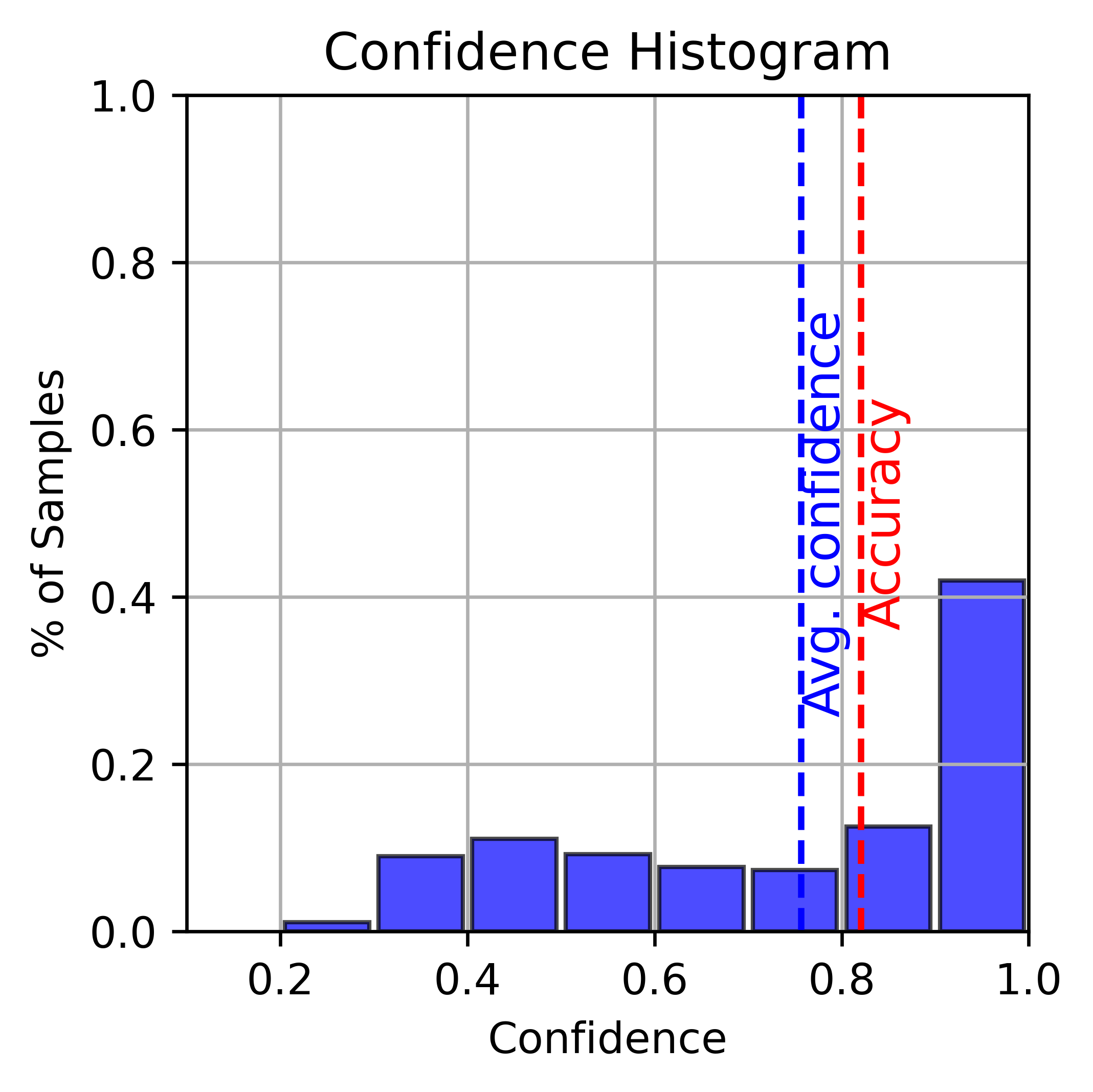}
    }\hspace{-1em} 
    \subfloat{
        \includegraphics[width=0.23\textwidth]{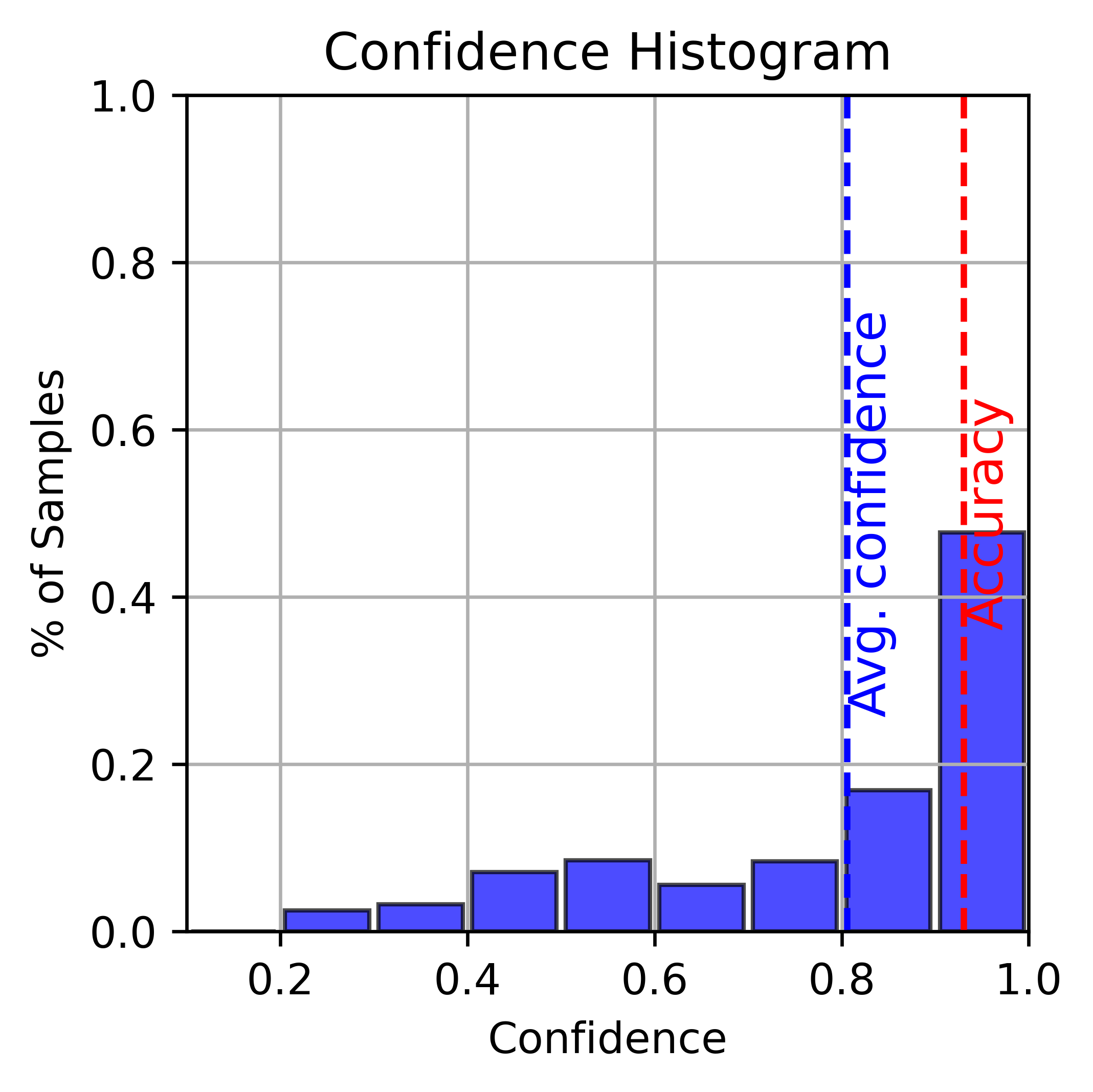}
    }\hspace{-1em} 
    \subfloat{
        \includegraphics[width=0.23\textwidth]{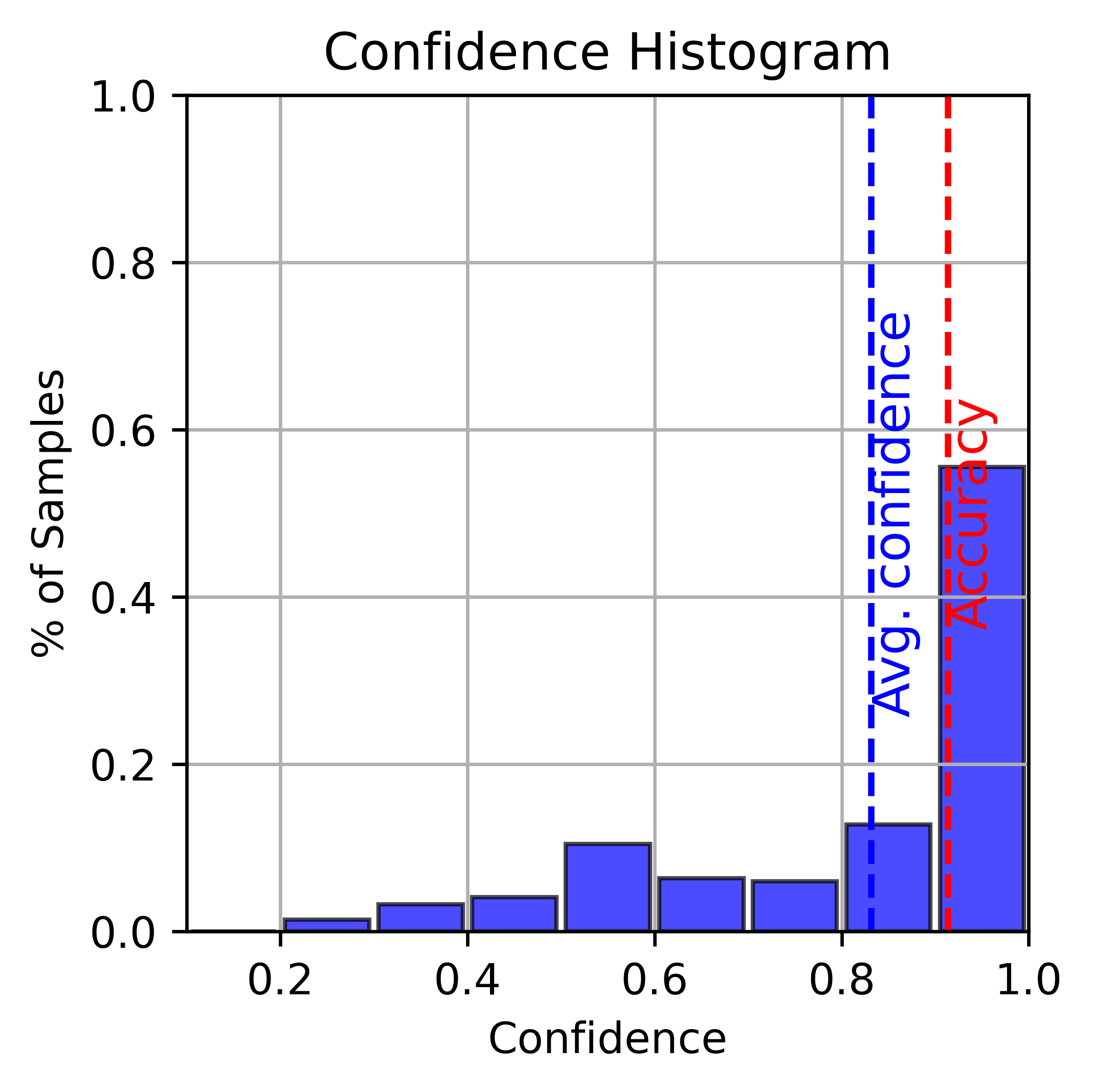}
    }
    \\[-1. em] 
    
    \subfloat[(a) Transfer Learning \label{fig:rel_1}]{
        \includegraphics[width=0.23\textwidth]{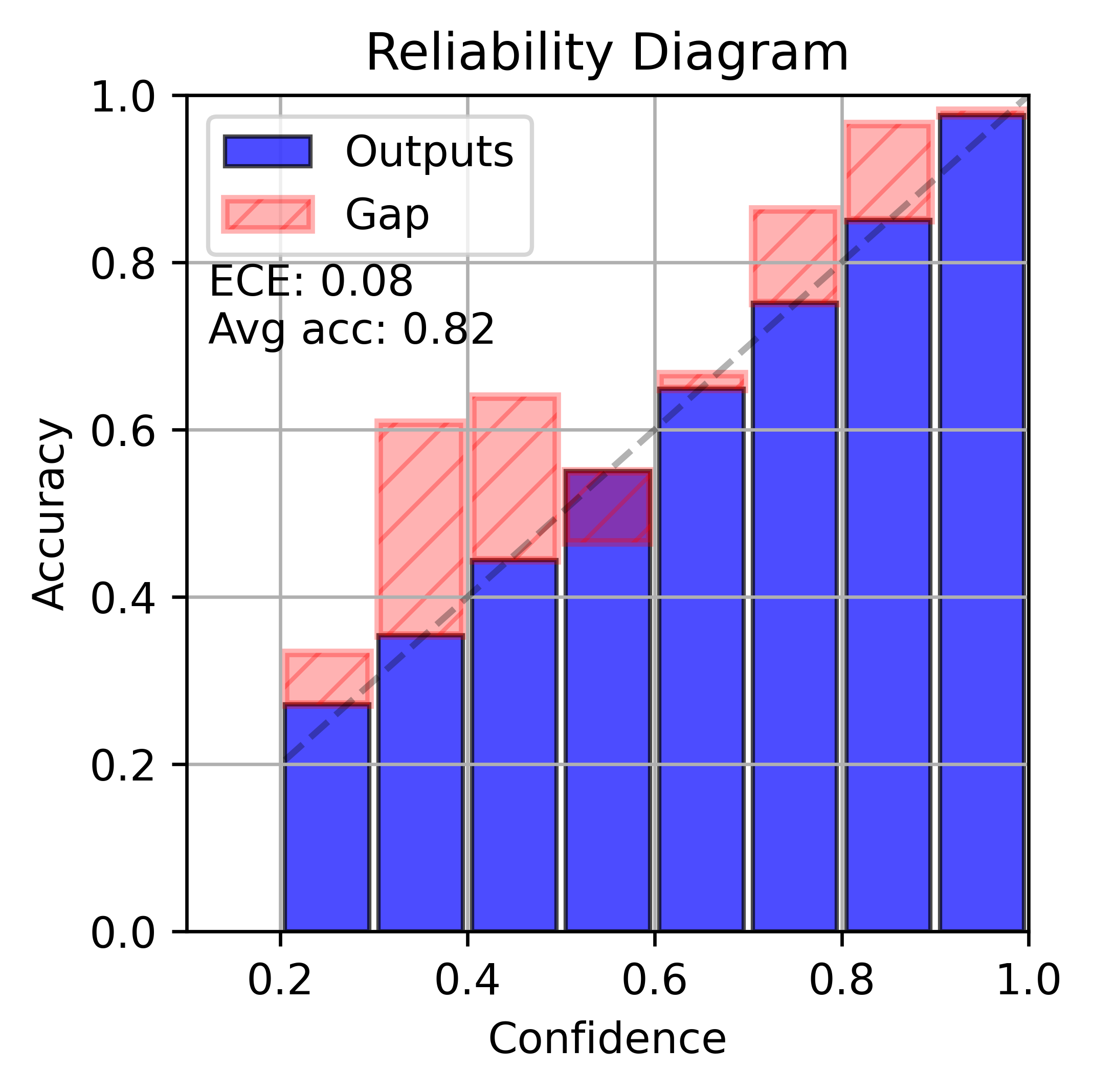}
    } \hspace{-1em} 
    \subfloat[(b) Model Retraining \label{fig:rel_2}]{
        \includegraphics[width=0.23\textwidth]{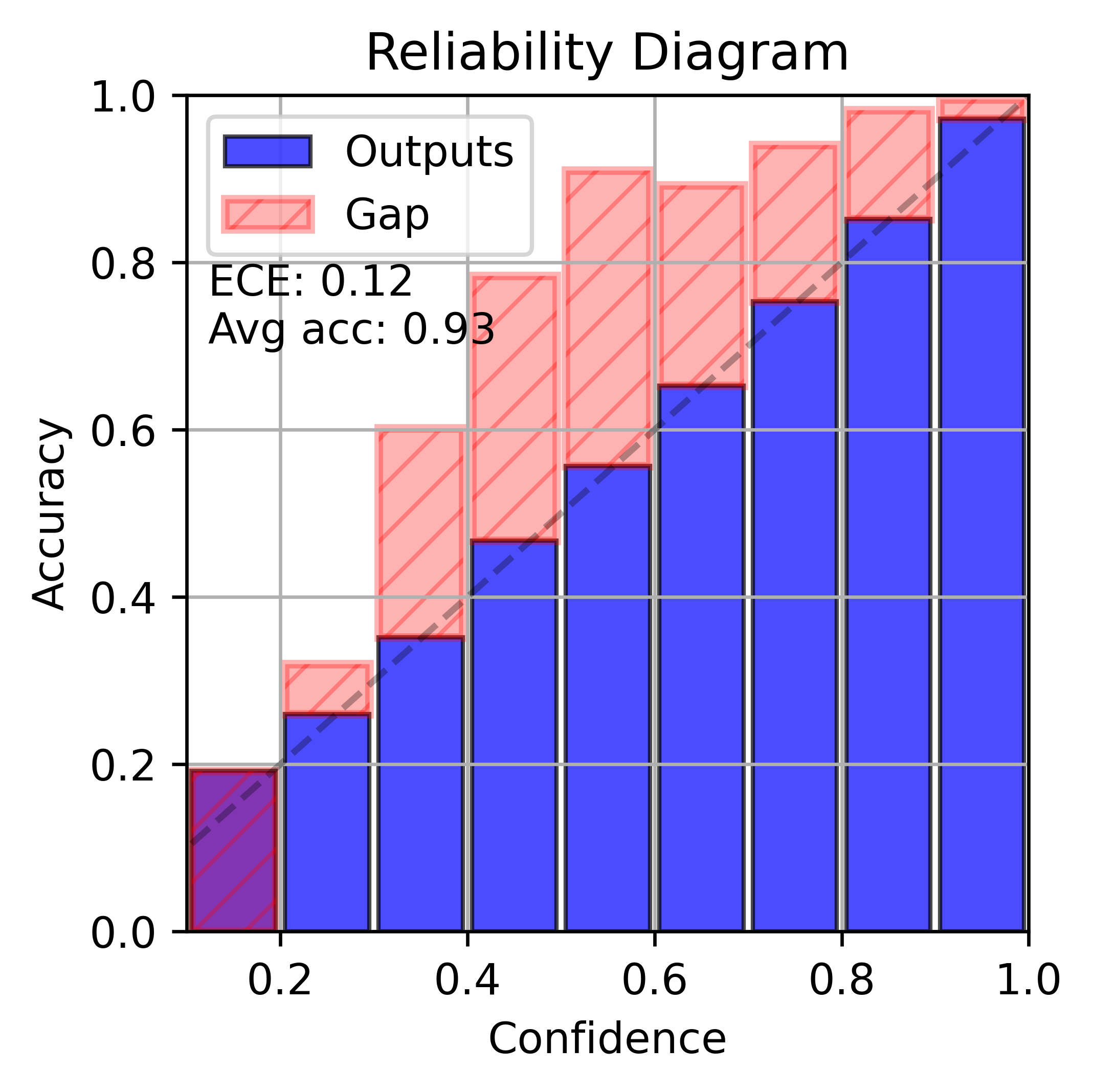}
    } \hspace{-1em} 
    \subfloat[(c) Poster.-aided Continual Learning \label{fig:rel_3}]{
        \includegraphics[width=0.23\textwidth]{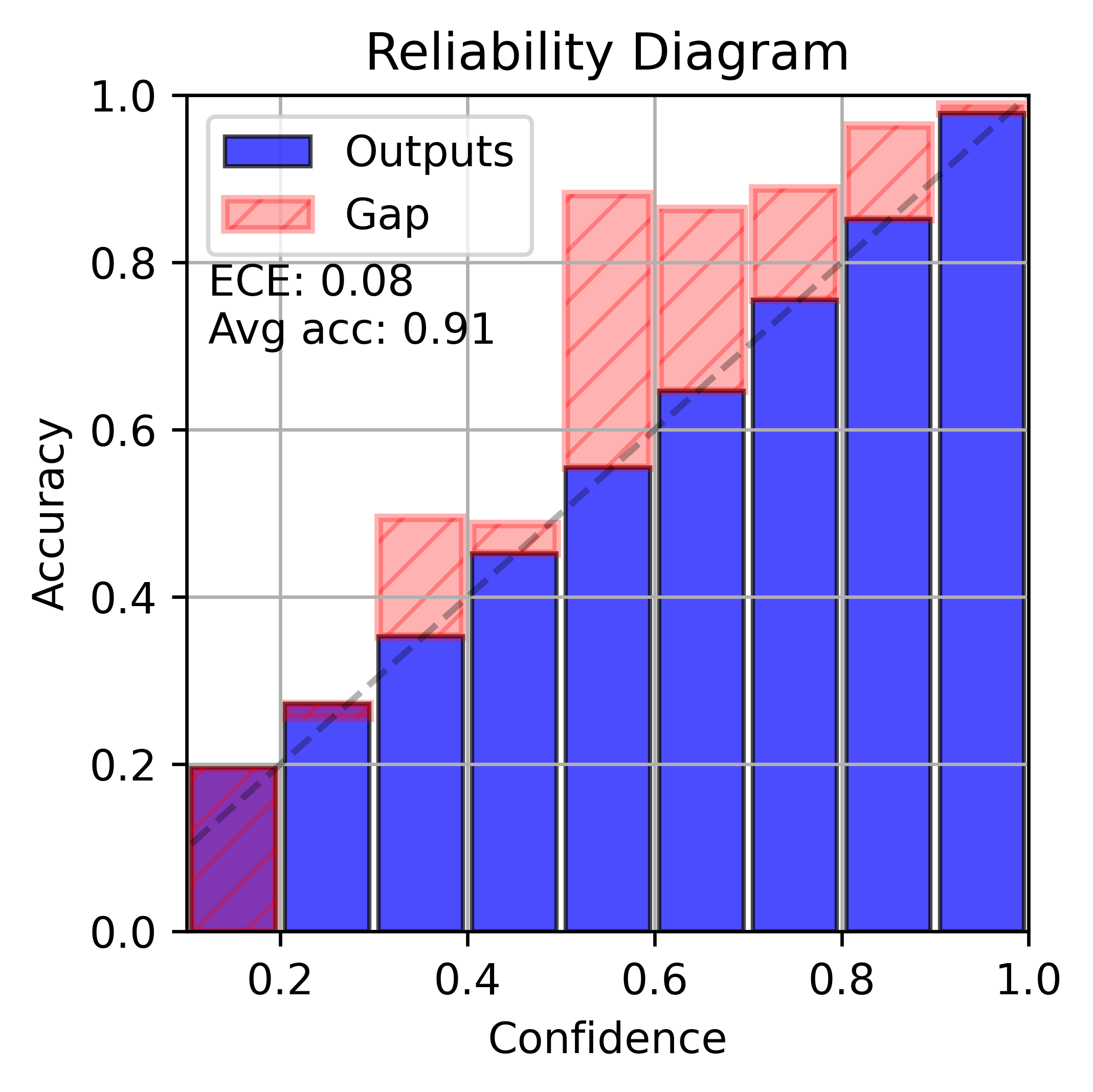}
    } 
    
    \caption{Confidence histograms (top) and reliability diagrams (bottom) for the \textit{Transfer Learning} (left), \textit{Model Retraining} (middle) and \textit{Posterior-aided Continual Learning} (right). The blue color represents the model \textit{outputs} (confidence), whereas the red color represents the calibration \textit{gap}, which is the difference between accuracy and confidence for a given bin.}
    \label{fig:results}
\end{figure*}

We evaluate the proposed approach to continual BFL over a network of $N=10$ nodes. For each day $d$ the number of samples on each node is set to $50$ randomly distributed range-azimuth maps from the set of samples $\mathcal{D}^d$ to form $\mathcal{D}_n^d$. At each iteration, all the nodes transmit the newly updated parameters to the PS where the aggregation is performed as the weighted average over the nodes. For all of the cases below, we perform the training for $T=100$ iteration with $T_b = 50$. Each node performs the SGD training with the learning rate $\eta = 10^{-4}$. The prior of our approach for $d=1$ is set to $p(\boldsymbol{\theta}^1) \sim \mathcal{N}(\mathbf{0}_{N_p}, \text{I}_{N_p})$, while for the following days the prior is set to $p(\boldsymbol{\theta}^d) \sim \mathcal{N}(\boldsymbol{\mu}^{d-1}, \mathbf{\Sigma}^{d-1})$ by computing the mean  and standard  deviation $\mathbf{\sigma}$ for each parameter $ \boldsymbol{\mu}^d = [\mu^d_1, ..., \mu^d_{N_p}]$, $\mathbf{\Sigma}^d = \operatorname{diag}([(\sigma_1^{d})^2, \dots, (\sigma_{N_p}^{d})^2])$ from the samples of the posterior for $d-1$. The results are presented in Fig. \ref{fig:results}.

We evaluate our approach in terms of validation accuracy, convergence speed, and reliability. For convergence speed, we select a specific accuracy threshold and analyze the number of iterations required to reach the set values. The performance metric chosen for assessing the reliability is the widely adopted Expected Calibration Error (ECE) \cite{callibr}. This metric quantifies the mismatch between the accuracy and the confidence of the model. First the confidence range $[0,1]$ is split into $J$  non-overlapping bins such that $\mathcal{D} = \{\mathcal{B}_j\}_{j=1}^J$. For each bin $B_j$ we compute the accuracy and the confidence respectively:

\begin{equation}
    \text{acc}(B_j) = \frac{1}{|B_j|} \sum_{i \in B_j} \mathbf{1} (\hat{y}_i = y_i),
\end{equation}

\begin{equation}
    \text{conf}(B_j) = \frac{1}{|B_j|} \sum_{i \in B_j} \hat{p}_i,
\end{equation}
where $|B_m|$ is the number of samples $j$-th confidence bin, $\hat{y}_i$ is the predicted label, $y_i$ is the true label, and $\hat{p}_i$ is the confidence score associated with the prediction. 
We define the ECE as the weighted average of the absolute differences between the accuracy and the confidence for each bin, namely:
\begin{equation}
    \text{ECE} = \sum_{j=1}^{J} \frac{|B_J|}{|D|} \left| \text{acc}(B_j) - \text{conf}(B_j) \right|.
\end{equation}

A lower ECE value indicates better model calibration, meaning the predicted confidence aligns more closely with the actual accuracy.

In what follows, we compare three different setups:
\begin{itemize}
    \item \textbf{Transfer Learning (TL)}. Corresponds to a case when the BFL model is trained on $\mathcal{D}^1$, and applied to all the subsequent days without further update (from day $d=1$ to $d=3$);
    
    \item \textbf{Model Retraining (Retr.)}. In this case we train a new model from scratch for each day's data $\mathcal{D}^d$ with $p(\boldsymbol{\theta}^{d}) \sim \mathcal{N}(\mathbf{0}_{N_p}, \mathbf{I}_{N_p})$ and evaluate it on the respected day's data;
    
    \item \textbf{Posterior-aided Continual Learning (P-CL)}. As described in Sect. \ref{CBFL}, we retrain the model in the days $d>1$ by replacing the prior with the past posterior obtained according to (\ref{eq:bayes}) using the sample mean and covariance matrix obtained from posterior samples of $d-1$.

\end{itemize}


We highlight the need for continuous training by providing the performance of \textit{TL} for all the days, as shown in the upper part of Table \ref{tab:res}. First, we compared the approaches in terms of accuracy and reliability, as shown in Fig. \ref{fig:cont_learn_fig}. \textit{TL} serves as a starting point providing the lowest accuracy over the days. \textit{Retr.} case manages to compensate for the accuracy, yet it introduces a larger \textit{ECE} compared to the first approach, as also seen from the gap between the accuracy and average confidence on the confidence histogram. The proposed \textit{P-CL} approach manages to achieve similar performance to the \textit{Retr.}, yet with increased reliability by having a lower ECE. Second, we contrast the \textit{Retr.} and \textit{P-CL} approaches in terms of convergence speed, namely the required number of iterations to achieve a target accuracy level of $85 \%$ per day, as shown in the bottom part of Table \ref{tab:res}. The threshold was set based on the based on the accuracies from Fig. \ref{fig:results}. The results indicate the benefit of leveraging the previous posterior for the optimization process by reducing the number of iterations needed and the communication overhead by almost $40 \%$ for day 2, $34 \%$ for day 3 and $50 \%$ for the whole simulation, with respect to the $\textit{Retr.}$ case. It is important to highlight, that the number of iterations for the first day is identical for the approaches as the prior is identical for both cases.

\begin{table}[t]
\centering
\begin{tabular}{c|ccc|}
\cline{2-4}
\multicolumn{1}{l|}{\textbf{}}                 & \multicolumn{1}{c|}{Day 1} & \multicolumn{1}{c|}{Day 2} & Day 3 \\ \hline
\multicolumn{1}{|c|}{\textit{\textbf{Setup}}}           & \multicolumn{3}{c|}{\textbf{Acc., \%}}                          \\ \hline
\multicolumn{1}{|c|}{\textbf{TL}} & \multicolumn{1}{c|}{95.5} & \multicolumn{1}{c|}{79.9} & 72.7 \\ \hline
\multicolumn{1}{|c|}{\cellcolor[HTML]{C0C0C0}} & \multicolumn{3}{c|}{\textbf{Num. Iter. \textit{(Acc.=85\%)}}}            \\ \hline
\multicolumn{1}{|c|}{\textbf{Retr.}}           & \multicolumn{1}{c|}{21}    & \multicolumn{1}{c|}{36}    & 47    \\ \hline
\multicolumn{1}{|c|}{\textbf{P-CL}}    & \multicolumn{1}{c|}{21}    & \multicolumn{1}{c|}{14 (-40\%)}    & 16 (-34\%)  \\ \hline
\end{tabular}
\caption{The accuracy of the model for each day using \textit{Transfer Learning} and the number of iterations per day to achieve the set accuracy.}
\label{tab:res}
\vspace{-0.1cm}
\end{table}

\section{Conclusions} \label{CONCL}
This paper discussed a variation of Bayesian Federated Learning (BFL) based on gradient-based Markov Chain Monte Carlo (MCMC) applied to a continual learning problem. We considered an industrial setup where a human operator is localized inside a robotized workplace and within predefined Regions of Interest (ROIs). In this setup, the data distribution obtained from radar sensors shifts over time, leading to a misalignment between daily observations. We treated the problem in a continual learning fashion, incorporating the posterior from previous training iterations to assist the learning process. We compared our approach against a transfer learning baseline using data from the first day and a full daily model retraining with a standard Gaussian prior. Performance was evaluated in terms of accuracy, convergence speed, and reliability (Expected Calibration Error). Results indicate that using the previous posterior as prior for continual learning not only improves the final performance and convergence speed but also enhances the model's reliability. This demonstrates the potential of Bayesian Federated Learning for continual adaptation, while maintaining calibration in non-abrupt dynamic environments, which is the main assimption of this work.

Future work will involve the exploration of more expressive priors, such as hierarchical Bayesian models or nonparametric approaches (e.g., Gaussian Processes) to further improve adaptability, and analysis of privacy constraints while retaining the benefits of Bayesian inference. Alternatively, exploration of privacy-preserving techniques in combination with our approach, such as differential privacy or homomorphic encryption, will be investigated for real-world deployments.

\bibliographystyle{IEEEtran}
\bibliography{bibliography}

\begin{thebibliography}{10}
\providecommand{\url}[1]{#1}
\csname url@samestyle\endcsname
\providecommand{\newblock}{\relax}
\providecommand{\bibinfo}[2]{#2}
\providecommand{\BIBentrySTDinterwordspacing}{\spaceskip=0pt\relax}
\providecommand{\BIBentryALTinterwordstretchfactor}{4}
\providecommand{\BIBentryALTinterwordspacing}{\spaceskip=\fontdimen2\font plus
\BIBentryALTinterwordstretchfactor\fontdimen3\font minus \fontdimen4\font\relax}
\providecommand{\BIBforeignlanguage}[2]{{%
\expandafter\ifx\csname l@#1\endcsname\relax
\typeout{** WARNING: IEEEtran.bst: No hyphenation pattern has been}%
\typeout{** loaded for the language `#1'. Using the pattern for}%
\typeout{** the default language instead.}%
\else
\language=\csname l@#1\endcsname
\fi
#2}}
\providecommand{\BIBdecl}{\relax}
\BIBdecl

\bibitem{FL_main}
\BIBentryALTinterwordspacing
H.~B. McMahan, E.~Moore, D.~Ramage, S.~Hampson, and B.~A. y~Arcas, ``Communication-efficient learning of deep networks from decentralized data,'' 2017. [Online]. Available: \url{https://arxiv.org/abs/1602.05629}
\BIBentrySTDinterwordspacing

\bibitem{FL_health}
U.~Milasheuski, L.~Barbieri, B.~C. Tedeschini, M.~Nicoli, and S.~Savazzi, ``On the impact of data heterogeneity in federated learning environments with application to healthcare networks,'' in \emph{2024 IEEE Conference on Artificial Intelligence (CAI)}, 2024, pp. 1017--1023.

\bibitem{FL_snarthome}
B.~Zhu, K.~Lu, and T.~Tao, ``A blockchain-based federated learning for smart homes,'' in \emph{2023 4th International Conference on Information Science, Parallel and Distributed Systems (ISPDS)}, 2023, pp. 689--693.

\bibitem{FL_auton}
\BIBentryALTinterwordspacing
A.~Nguyen, T.~Do, M.~Tran, B.~X. Nguyen, C.~Duong, T.~Phan, E.~Tjiputra, and Q.~D. Tran, ``Deep federated learning for autonomous driving,'' 2022. [Online]. Available: \url{https://arxiv.org/abs/2110.05754}
\BIBentrySTDinterwordspacing

\bibitem{FL_telco}
\BIBentryALTinterwordspacing
V.~T. Nguyen and R.~Beuran, ``Fedmse: Federated learning for iot network intrusion detection,'' 2024. [Online]. Available: \url{https://arxiv.org/abs/2410.14121}
\BIBentrySTDinterwordspacing

\bibitem{comm_mag}
S.~Savazzi, M.~Nicoli, M.~Bennis, S.~Kianoush, and L.~Barbieri, ``Opportunities of federated learning in connected, cooperative, and automated industrial systems,'' \emph{IEEE Communications Magazine}, vol.~59, no.~2, pp. 16--21, 2021.

\bibitem{BFL}
\BIBentryALTinterwordspacing
L.~Cao, H.~Chen, X.~Fan, J.~Gama, Y.-S. Ong, and V.~Kumar, ``Bayesian federated learning: A survey,'' 2023. [Online]. Available: \url{https://arxiv.org/abs/2304.13267}
\BIBentrySTDinterwordspacing

\bibitem{BI}
\BIBentryALTinterwordspacing
D.~J.~C. MacKay, ``Probable networks and plausible predictions - a review of practical bayesian methods for supervised neural networks,'' \emph{Network: Computation In Neural Systems}, vol.~6, pp. 469--505, 1995. [Online]. Available: \url{https://api.semanticscholar.org/CorpusID:14332165}
\BIBentrySTDinterwordspacing

\bibitem{VI}
\BIBentryALTinterwordspacing
M.~Ashman, T.~D. Bui, C.~V. Nguyen, S.~Markou, A.~Weller, S.~Swaroop, and R.~E. Turner, ``Partitioned variational inference: A framework for probabilistic federated learning,'' 2022. [Online]. Available: \url{https://arxiv.org/abs/2202.12275}
\BIBentrySTDinterwordspacing

\bibitem{VI2}
\BIBentryALTinterwordspacing
R.~Kassab and O.~Simeone, ``Federated generalized bayesian learning via distributed stein variational gradient descent,'' 2021. [Online]. Available: \url{https://arxiv.org/abs/2009.06419}
\BIBentrySTDinterwordspacing

\bibitem{MCMC}
S.~Ahn, B.~Shahbaba, and M.~Welling, ``Distributed stochastic gradient mcmc,'' in \emph{Proceedings of the 31st International Conference on International Conference on Machine Learning - Volume 32}, ser. ICML'14.\hskip 1em plus 0.5em minus 0.4em\relax JMLR.org, 2014, p. II–1044–II–1052.

\bibitem{SGLD}
M.~Welling and Y.~W. Teh, ``Bayesian learning via stochastic gradient langevin dynamics,'' in \emph{Proceedings of the 28th International Conference on International Conference on Machine Learning}, ser. ICML'11.\hskip 1em plus 0.5em minus 0.4em\relax Madison, WI, USA: Omnipress, 2011, p. 681–688.

\bibitem{wirelessSGLD}
D.~Liu and O.~Simeone, ``Wireless federated langevin monte carlo: Repurposing channel noise for bayesian sampling and privacy,'' \emph{IEEE Transactions on Wireless Communications}, vol.~22, no.~5, pp. 2946--2961, 2023.

\bibitem{MCMC3}
\BIBentryALTinterwordspacing
W.~Deng, Q.~Zhang, Y.-A. Ma, Z.~Song, and G.~Lin, ``On convergence of federated averaging langevin dynamics,'' 2023. [Online]. Available: \url{https://arxiv.org/abs/2112.05120}
\BIBentrySTDinterwordspacing

\bibitem{MCMC4}
\BIBentryALTinterwordspacing
M.~Gürbüzbalaban, X.~Gao, Y.~Hu, and L.~Zhu, ``Decentralized stochastic gradient langevin dynamics and hamiltonian monte carlo,'' 2021. [Online]. Available: \url{https://arxiv.org/abs/2007.00590}
\BIBentrySTDinterwordspacing

\bibitem{cont_health}
\BIBentryALTinterwordspacing
P.~Kumari, J.~Chauhan, A.~Bozorgpour, B.~Huang, R.~Azad, and D.~Merhof, ``Continual learning in medical image analysis: A comprehensive review of recent advancements and future prospects,'' 2024. [Online]. Available: \url{https://arxiv.org/abs/2312.17004}
\BIBentrySTDinterwordspacing

\bibitem{cont_iot}
\BIBentryALTinterwordspacing
J.~Chen, J.~He, F.~Chen, Z.~Lv, J.~Tang, W.~Li, Z.~Liu, H.~H. Yang, and G.~Han, ``Towards general industrial intelligence: A survey of continual large models in industrial iot,'' 2024. [Online]. Available: \url{https://arxiv.org/abs/2409.01207}
\BIBentrySTDinterwordspacing

\bibitem{CL}
\BIBentryALTinterwordspacing
L.~Wang, X.~Zhang, H.~Su, and J.~Zhu, ``A comprehensive survey of continual learning: Theory, method and application,'' 2024. [Online]. Available: \url{https://arxiv.org/abs/2302.00487}
\BIBentrySTDinterwordspacing

\bibitem{CBFL1}
\BIBentryALTinterwordspacing
H.~Zhang, L.~Yang, Z.~Qin, Q.~Wang, Y.~Han, Q.~Hu, and Y.~Deng, ``Variational federated continual learning,'' 2024. [Online]. Available: \url{https://openreview.net/forum?id=lzt60v45V4}
\BIBentrySTDinterwordspacing

\bibitem{CBFL2}
\BIBentryALTinterwordspacing
D.~Yao, S.~Li, Y.~Dai, Z.~Xu, S.~Hu, P.~Zhao, and L.~Sun, ``Variational bayes for federated continual learning,'' 2024. [Online]. Available: \url{https://arxiv.org/abs/2405.14291}
\BIBentrySTDinterwordspacing

\bibitem{Luca}
L.~Barbieri, S.~Savazzi, and M.~Nicoli, ``On the impact of model compression for bayesian federated learning: An analysis on healthcare data,'' \emph{IEEE Signal Processing Letters}, vol.~32, pp. 251--255, 2025.

\bibitem{HR-INT}
\BIBentryALTinterwordspacing
S.~Hjorth and D.~Chrysostomou, ``Human–robot collaboration in industrial environments: A literature review on non-destructive disassembly,'' \emph{Robotics and Computer-Integrated Manufacturing}, vol.~73, p. 102208, 2022. [Online]. Available: \url{https://www.sciencedirect.com/science/article/pii/S0736584521000910}
\BIBentrySTDinterwordspacing

\bibitem{RH-SAFETY}
S.~Kianoush, S.~Savazzi, M.~Beschi, S.~Sigg, and V.~Rampa, ``A multisensory edge-cloud platform for opportunistic radio sensing in cobot environments,'' \emph{IEEE Internet of Things Journal}, vol.~8, no.~2, pp. 1154--1168, 2021.

\bibitem{dataset}
\BIBentryALTinterwordspacing
S.~Savazzi, ``Federated learning: mmwave mimo radar dataset for testing,'' 2020. [Online]. Available: \url{https://dx.doi.org/10.21227/0wmc-hq36}
\BIBentrySTDinterwordspacing

\bibitem{LeNet}
Y.~Lecun, L.~Bottou, Y.~Bengio, and P.~Haffner, ``Gradient-based learning applied to document recognition,'' \emph{Proceedings of the IEEE}, vol.~86, no.~11, pp. 2278--2324, 1998.

\bibitem{callibr}
\BIBentryALTinterwordspacing
C.~Guo, G.~Pleiss, Y.~Sun, and K.~Q. Weinberger, ``On calibration of modern neural networks,'' 2017. [Online]. Available: \url{https://arxiv.org/abs/1706.04599}
\BIBentrySTDinterwordspacing

\end{thebibliography}


\end{document}